%% file: main.tex
\documentclass[a4paper,11pt]{article}
\pdfoutput=1 

\usepackage{jinstpub} 

\usepackage{graphicx}
\usepackage{wrapfig}
\usepackage{subcaption}

\usepackage[acronym,nonumberlist,nopostdot,toc]{glossaries}
\glsdisablehyper
\usepackage{glossaries-prefix}
\loadglsentries[type=\acronymtype]{acronyms}

\usepackage{tikz}
\usetikzlibrary{shapes.geometric, arrows}
\usetikzlibrary{shapes.multipart}

\title{\boldmath Mitigating multiple single-event upsets during deep neural network inference using fault-aware training}

\author[a,b,c,1]{Toon~Vinck,\note{Corresponding author.}}
\author[a,b]{Naïn~Jonckers,}
\author[a]{Gert~Dekkers,}
\author[b]{Jeffrey~Prinzie}
\author[c,d]{and Peter~Karsmakers}

\affiliation[a]{Magics Technologies,\\Cipalstraat 3, 2440 Geel, Belgium}
\affiliation[b]{ESAT-ADVISE, KU Leuven,\\Kleinhoefstraat 4, 2440 Geel, Belgium}
\affiliation[c]{Dept. of Computer Science, Leuven.AI, KU Leuven\\Celestijnenlaan 200a, 3001 Leuven, Belgium}
\affiliation[d]{MPRO, Flanders Make,\\Gaston Geenslaan 8, 3001 Leuven, Belgium}

\emailAdd{toon.vinck@magics.tech}

\abstract{Deep neural networks (DNNs) are increasingly used in safety-critical applications. Reliable fault analysis and mitigation are essential to ensure their functionality in harsh environments that contain high radiation levels. This study analyses the impact of multiple single-bit single-event upsets in DNNs by performing fault injection at the level of a DNN model. Additionally, a \gls{fat} methodology is proposed that improves the DNNs' robustness to faults without any modification to the hardware. Experimental results show that the \gls{fat} methodology improves the tolerance to faults up to a factor 3.}


\proceeding{Topical Workshop on Electronics for Particle Physics\\
  30 September -- 4 October 2024\\
  Glasgow, Scotland, United Kingdom}

\begin{document}
\maketitle
\flushbottom

\section{Introduction}
\label{sec:intro}

Over the past decade \glspl{dnn} have made remarkable advancements in terms of performance. Special hardware accelerators, designed to optimise the execution of these algorithms, have played a crucial role in this progress. However, their robustness remains a concern for safety-critical operations, especially when deployed in environments that contain high levels of radiation. In these harsh environments, accelerators are susceptible to \glspl{seu}, which can lead to bit-flips causing numerical errors or even a system crash. \par

One approach to mitigate \glspl{seu} is \gls{rhbd}. As this implies adding redundancy to the hardware, it comes with significant overheads in terms of energy consumption, surface area and optionally latency. As \glspl{dnn} inherently possess redundancy, the requirements for \gls{rhbd} might be relaxed and hence the latter overhead might be reduced. \par

To assess the robustness of \glspl{dnn} when subjected to random bit-flips, which will be referred to as \textit{faults} in the remainder of the paper, \gls{fi} can be employed at different abstraction levels. While low-level \gls{fi} is known to be more realistic, it requires access to the underlying hardware.
This work uses \gls{fi} in the high-level deep learning library which is, although known to be less accurate, a more accessible and faster alternative that is independent of the used hardware implementation \cite{vinck2024understanding}.\par

Previous studies have explored the impact of such faults in \glspl{dnn}. Li et al. \cite{li2017understanding} investigated the impact of faults in \glspl{dnn} using different data types, values and architectures. 
Goldstein et al. \cite{goldstein2020reliability} further extended this line of research by examining the impact of faults on quantised and compressed \glspl{dnn}, which are typically employed when resources are limited.
These studies show the probability of an erroneous output given a fault during a \gls{dnn} inference. However, they do not consider the impact of different amounts of injected faults during a single inference. Therefore, this research investigates the relationship between the performance and the number of injected faults.\par 

Next to a proper fault analysis this paper also proposes a fault mitigation strategy by means of adapting the parameters of a \gls{dnn} similar to \cite{zahid2020fat}. This strategy, known as \acrfull{fat}, is based on the hypothesis that \glspl{dnn} can be trained to be aware of these faults.
\par
This paper starts with the design of a \gls{fi} tool designed on top of the PyTorch \gls{dnn} library \cite{paszke2017automatic} in section \ref{sec:fi_methodologies}. This tool is used in section \ref{sec:robustness_analysis} to conduct \gls{fi} experiments on two conventionally trained \glspl{dnn}. Section \ref{sec:fat} presents the \gls{fat} methodology employed to train the same \glspl{dnn} and compares the robustness of these \gls{fat} models to that of conventionally trained models. Finally, section \ref{sec:conclusion} discusses the implications of multiple faults in \glspl{dnn}.

\section{Fault injection methodology}
\label{sec:fi_methodologies}
    
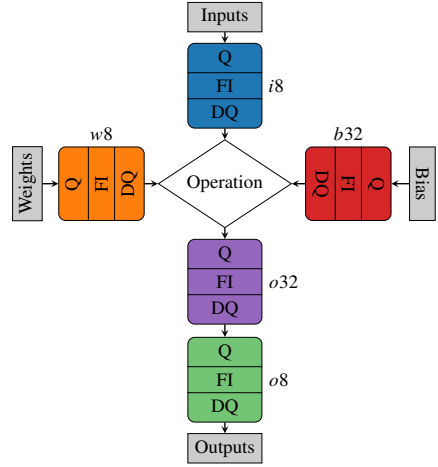
\begin{wrapfigure}{r}{0.5\textwidth}
    \centering
            \resizebox{0.48\columnwidth}{!}
            {\input{images/fanns_fi_flow}}
        \caption{Representation of the \gls{fi} tool applied to one quantised DNN layer \cite{vinck2024understanding}.}
        \label{fig:repr_Qlayer}
\end{wrapfigure}
        \glspl{seu} which occur in an accelerator during \gls{dnn} inference may cause bit-flips which can lead to an erroneous output. To analyse the susceptibility to these faults, a \gls{fi} tool was developed.
        This tool aims to simulate multiple \glspl{sbu} within a \gls{dnn} accelerator, \glspl{mbu} caused by one ionising particle are ignored. \par
        
        A \gls{dnn} accelerator is comprised of a data path, responsible for the numerical operations, and a control path, responsible for orchestrating these operations. As most of the surface area is attributed to the data path \cite{chen2019eyeriss, jouppi2018motivation}, this paper only focusses on faults in the data path, the control path is assumed to be protected in hardware.\par
        
        Given the resource constraints typically encountered in harsh radiation environments, the \gls{fi} tool is build on top of a framework that uses quantised \gls{dnn} accelerator arithmetic as described by Nagel et al. \cite{nagel2021white}. This setup integrates in the PyTorch graph to make sure it does not interfere with the standard PyTorch execution.\par
        
        Figure \ref{fig:repr_Qlayer} shows a flowchart of the \gls{fi} tool applied to one layer of the \gls{dnn} in this setup. The operation node is the core of the flowchart, which executes a specific type of layer, such as a convolution or \gls{fc} layer. These operations are inherently executed using floating point data types. In order to generate an output tensor, these operations take into account input, weight and bias tensors. To simulate the quantised arithmetic, all these tensors are \textit{fake-quantised}.\par
        
        Equation \ref{eq:sym_quant} shows the symmetric quantisation that occurs in two steps. In the Q (quantise) step, the floating point tensor values $x_{f32}$ are quantised to integers $x_{int}$ using a floating point scale-factor $s$. The clamp operation is determined by the bit-width $b$ of the integers. In the DQ (dequantise) step, the integers $x_{int}$ are dequantised back to fake-quantised floating point values $\hat{x}_{f32}$. 
        \begin{equation}
            \hat{x}_{f32} = s \cdot x_{int} = s \cdot \text{clamp} \left( \biggl \lfloor \frac{x_{f32}}{s} \biggl \rceil, -2^{b-1}, 2^{b-1}-1\right)
            \label{eq:sym_quant}
        \end{equation}
        These steps are executed in five different places in the flowchart. Based on \cite{nagel2021white} 8-bit integers were used for the inputs ($i8$) and weights ($w8$). 32-bit integers were used for the bias ($b32$) and accumulator ($o32$), after which the output ($o8$) is quantised back to 8-bit integers. The five quantisation nodes are called modules.\par
        
     To simulate \glspl{seu} in the \gls{dnn}, a \gls{fi} node that can perform bit-flips to the quantised tensor, was inserted in each module between the Q and DQ nodes. The overall \gls{fi} process is now conducted in two steps. First, before each forward pass of the \gls{dnn}, a predetermined number of bits is selected across the entire \gls{dnn} network. Thereafter, during the forward pass, the selected bits are flipped by the corresponding \gls{fi} nodes in the relevant modules. Note that these modules are placed in the PyTorch \gls{dnn}-graph such that faults are injected in the correct \gls{dnn}-state. This method provides the flexibility to inject faults into tensors that are only temporarily available, such as inputs and outputs in intermediate \gls{dnn} layers.

\section{Robustness analysis}
\label{sec:robustness_analysis}
To understand the robustness of the \gls{dnn} to multiple faults, two \gls{fi} experiments on two quantised \gls{dnn} models were conducted using publicly available datasets containing small images which are separated in 10 classes:
\begin{enumerate}
    \item \textbf{CCDF:} A \gls{cnn} with two convolutional layers, one dropout layer and one \gls{fc} layer trained on the MNIST handwritten digit dataset \cite{lecun2010mnist}. 
    \item \textbf{MobileNetV2:} A \gls{cnn} designed in \cite{sandler2018mobilenetv2}, trained on the CIFAR10 small colored image dataset \cite{krizhevsky2009learning}.
\end{enumerate}
These experiments were repeated three times to create a confidence interval of 95\%. \par
\begin{figure}[h]
\centering 
\includegraphics[width=.47\textwidth]{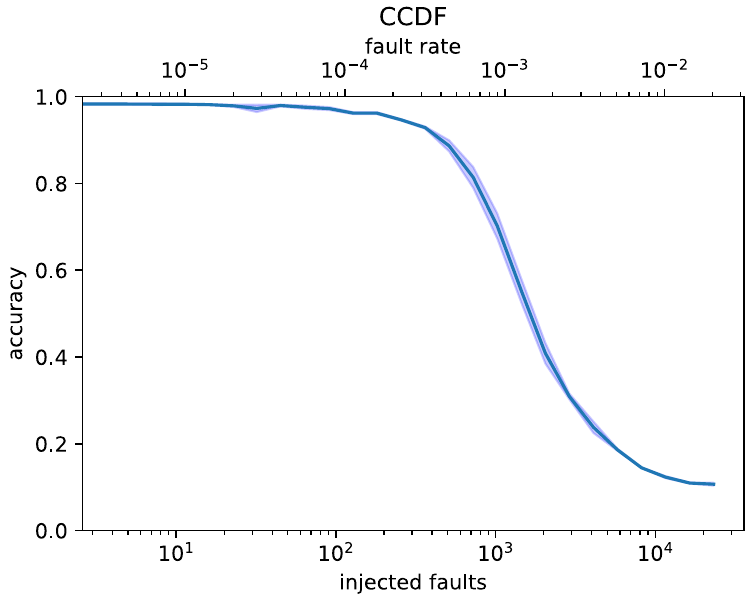}
\qquad
\includegraphics[width=.47\textwidth,]{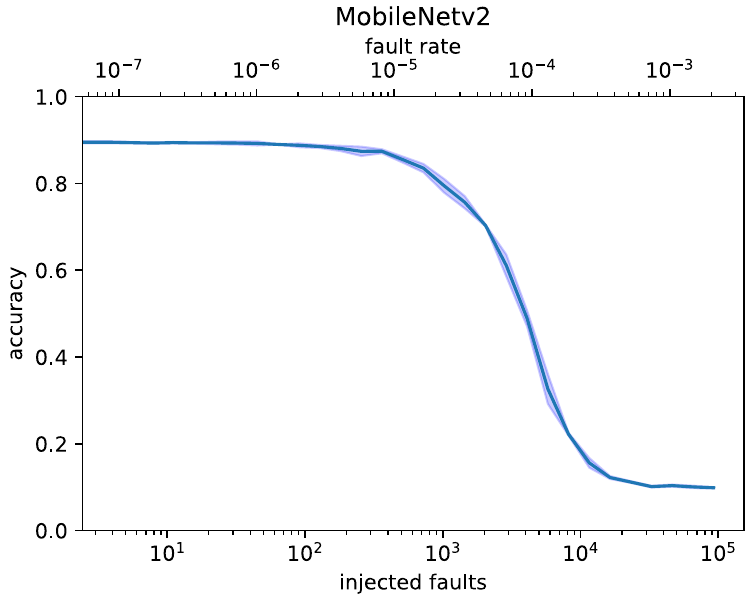}
\caption{\label{fig:ref_models} Relation between accuracy and the number of faults and fault rate, for the CCDF and MobileNetV2 experiment. The error bands represent a 95\% confidence interval.}
\end{figure}

The goal of the first experiment is to assess the overall robustness of the above models.
Utilising the \gls{fi} tool, faults were injected at random locations during the test phase of the \gls{dnn}, as explained in section \ref{sec:fi_methodologies}. The number of faults per inference was gradually increased as visualised in the x-axis of the plots in figure \ref{fig:ref_models}. For every number of injected faults, faults were injected across the entire test set of the corresponding dataset.\par

The results of this experiment, visualised in the plots in figure \ref{fig:ref_models}, show the relationship between the accuracy and the number of injected faults as well as the fault rate. The relationship between fault rate and the number of injected faults is given by equation \ref{eq:fault_rate}.
\begin{equation}
    \text{Fault rate} = \frac{\text{Injected faults per inference}}{\text{Amount vulnerable bits in one forward pass}}
    \label{eq:fault_rate}
\end{equation}
For both architectures, an initial increase in the number of faults, results in a decreased accuracy. As the number of faults continuous to rise, the accuracy eventually saturates at 10\%, which can be considered as random guessing in the employed classification datasets with 10 classes.
\par

\begin{figure}[h]
\centering 
\includegraphics[width=.47\textwidth]{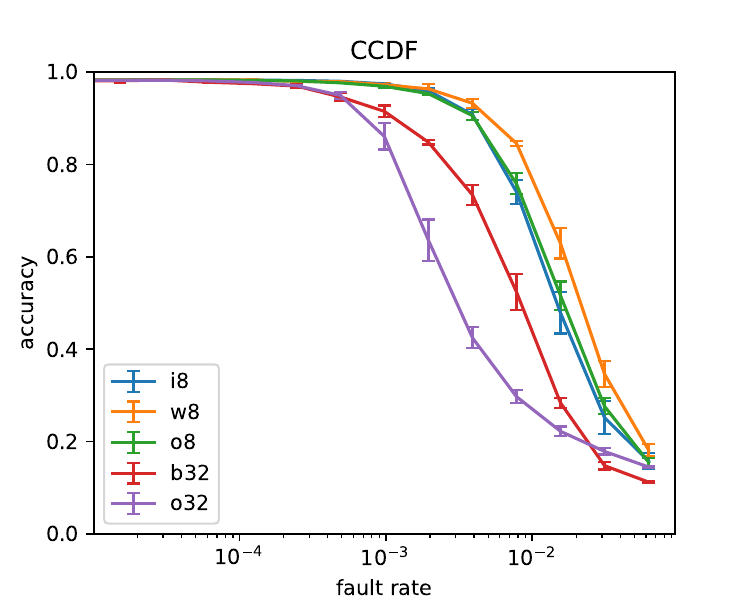}
\qquad
\includegraphics[width=.47\textwidth,]{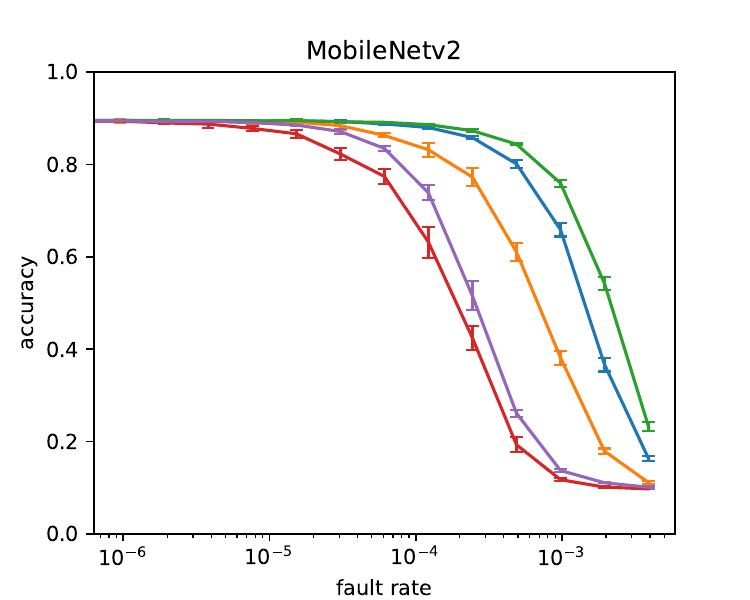}
\caption{\label{fig:one_module} Accuracy and fault rate relationship for the CCDF and MobileNetV2 experiment when faults are injected in only one module at a time. The error bars represent a 95\% confidence interval.}
\end{figure}

In a second experiment, the robustness of different components within the \gls{dnn} is analysed by injecting faults into one module at a time. Since each module contains a different number of bits, only the fault rate is considered. The results of this experiment are visualised in the plots in figure \ref{fig:one_module}. When comparing the two models, there is no consistent pattern regarding the sensitivity of the modules. However, it can be observed that the modules operating 32-bit data (the bias and accumulator) are more sensitive to faults than the 8-bit modules. 
This increased sensitivity can be attributed to the fact that the 32-bit modules usually have a high dynamic range relative to the stored data. Many of the \glspl{msb} remain unused. When bit-flips occur in these \glspl{msb}, they cause faults with disproportionately large magnitude, leading to a larger impact on the output of the \gls{dnn}.

\section{Fault aware training}
\label{sec:fat}
By injecting faults during the training phase of the \gls{dnn} models, the \gls{fat} technique aims to improve the robustness to these faults.
In order to evaluate the performance of the \gls{fat} methodology, a comparative experiment between \gls{fat} models and conventional models is designed.\par

As a reference, the first experiment outlined in section \ref{sec:robustness_analysis}, where the CCDF and MobileNetV2 models were conventionally trained and subsequently tested with an increasing number of injected faults, was repeated. To evaluate the effectiveness of \gls{fat}, the same models were trained using the \gls{fat} methodology, where an equal number of faults was injected during both the training and test phase. The numbers of faults used during \gls{fat} were identical to those in the reference experiment.
\par
However, a limitation arose when injecting faults into the 32-bit modules during training. During the training phase, the potential large magnitudes of faults in these modules would lead to an unstable adaptation of the scale factor $s$ introduced in equation \ref{eq:sym_quant}. This instability occurs because $s$ is dynamically calibrated during the training phase using the BatchNorm-based quantisation range settings \cite{nagel2021white}. The dynamic nature of this process makes it difficult to maintain stable quantisation under numerically large faults. This along with the relatively high impact of faults in the 32-bit modules motivates our choice to simulate the 32-bit modules to be protected in hardware. This is achieved by disabling the FI nodes in these modules.\par

\begin{figure}[t]
\centering
\includegraphics[width=.47\textwidth]{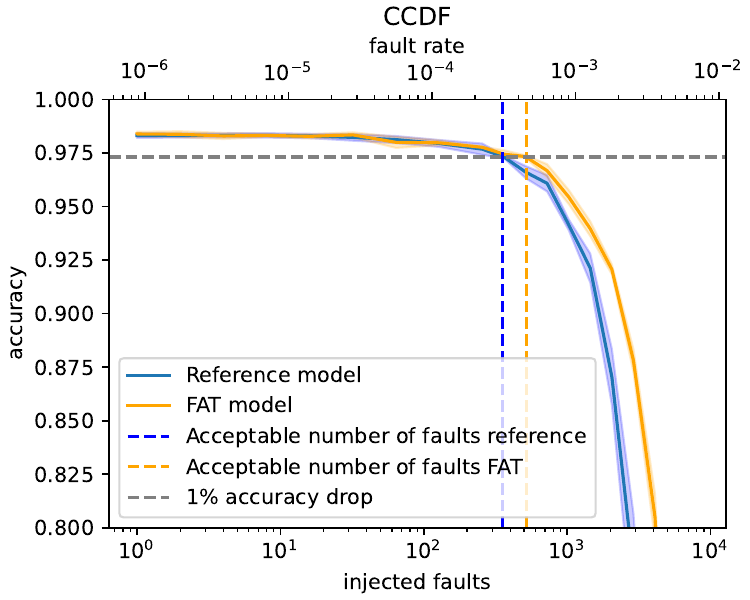}
\qquad
\includegraphics[width=.47\textwidth]{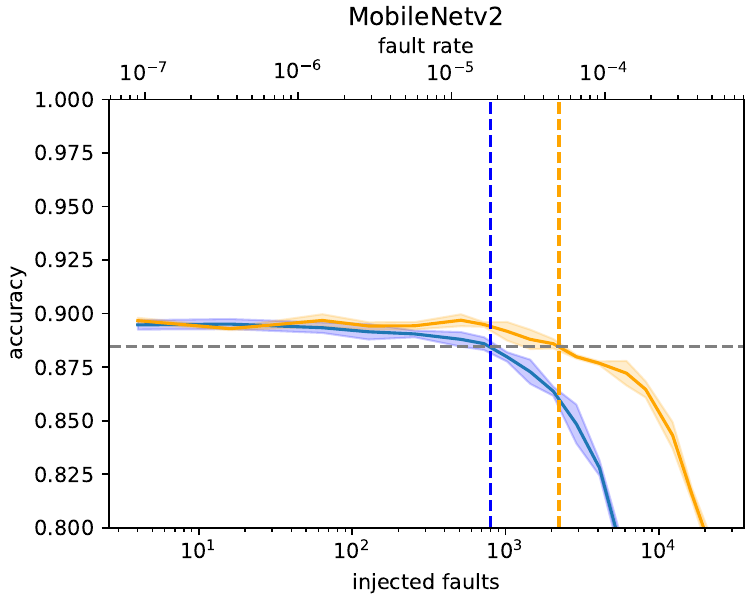}

\caption{\label{fig:FAT_exp} Comparison between the robustness of the \gls{fat} model and reference model. The error bands represent a 95\% confidence interval.}
\end{figure}

The results of the FAT experiment are visualised in the plots in figure \ref{fig:FAT_exp}. In absence of faults, the accuracy of both the \gls{fat} and reference models is identical. When a 1\% reduction in accuracy is tolerated, the \gls{fat} CCDF model can withstand more than 50\% more faults than the reference model, while the \gls{fat} MobileNetV2 model can handle 3 times more faults than the reference model. These results indicate that \gls{fat} can improve the tolerance of \glspl{dnn} to multiple faults. However, depending on the size of the model and the complexity of the dataset different gains can be expected.

\section{Conclusion}
\label{sec:conclusion}
This paper provided an experiment that analyses the impact of multiple \glspl{seu} on the performance of two quantised \glspl{dnn}. While \glspl{dnn} inherently exhibit a certain tolerance to these faults, the results show that the sensitivity to faults varies across different components within these \glspl{dnn}. Specifically, components with a high dynamic range relative to their data, such as the accumulator, are significantly more vulnerable to faults. To further enhance the fault robustness, a \gls{fat} methodology was proposed. A comparative experiment between \gls{fat} models and the original models showed that the \gls{fat} models can withstand up to 3 times more faults, suggesting that \glspl{dnn} can be designed to tolerate a significant number of faults.

\bibliographystyle{JHEP}
\bibliography{bibliography}






\end{document}

%% file: acronyms.tex
\newacronym{cnn}{CNN}{convolutional neural network}
\newacronym{dnn}{DNN}{deep neural network}
\newacronym{fat}{FAT}{fault-aware training}
\newacronym{fi}{FI}{fault injection}
\newacronym{fc}{FC}{fully connected}
\newacronym{mbu}{MBU}{multi-bit upset}
\newacronym{msb}{MSB}{most significant bit}
\newacronym{rhbd}{RHBD}{radiation hardening by design}
\newacronym{rtl}{RTL}{register-transfer level}
\newacronym{sbu}{SBU}{single-bit upset}
\newacronym{seu}{SEU}{single-event upset}

%% file: images/fanns_fi_flow.tex
{
\definecolor{blue0}{HTML}{3182BD}
\definecolor{blue1}{HTML}{1F77B4}
\definecolor{orange0}{HTML}{E6550D}
\definecolor{orange1}{HTML}{FF7F0E}
\definecolor{green0}{HTML}{31A354}
\definecolor{green1}{HTML}{74C476}
\definecolor{green2}{HTML}{A1D99B}
\definecolor{green3}{HTML}{C7E9C0}
\definecolor{purple0}{HTML}{756BB1}
\definecolor{purple1}{HTML}{9467BD}
\definecolor{grey0}{HTML}{636363}
\definecolor{grey1}{HTML}{969696}
\definecolor{red1}{HTML}{D62728}
\tikzstyle{io} = [rectangle, node distance=1.5cm, fill=white!80!black, draw, anchor=center, minimum width=1.5cm]
\tikzstyle{module} = [rectangle split, rectangle split parts=3, rounded corners, minimum width=1.5cm, minimum height = 0.5cm, text centered, align=center, draw=black]
\tikzstyle{operation} = [diamond, aspect=1.5, text centered, draw=black]
\tikzstyle{arrow} = [thick, ->, >=stealth]
\tikzstyle{i8_sty}=[draw, fill=blue1]
\tikzstyle{w8_sty}=[draw, fill=orange1]
\tikzstyle{o32_sty}=[draw, fill=purple1]
\tikzstyle{b32_sty}=[draw, fill=red1]
\tikzstyle{o8_sty}=[draw, fill=green1]
\tikzstyle{nlf8_sty}=[draw, fill=grey1]
\newcommand\drawmodule[7]{
    \node (#1) [module, #5_sty, #2, #3]{Q
        \nodepart{two} FI#7
        \nodepart{three} DQ};
    \ifthenelse{\equal{#6}{e}}{%
        \node (#1-label) at (#1.east) [anchor=west]{$#5$};
    }{\ifthenelse{\equal{#6}{e-s}}{
        \node (#1-label) at (#1.east) [anchor=south]{$#5$};
    }{\ifthenelse{\equal{#6}{w-s}}{
        \node (#1-label) at (#1.west) [anchor=south]{$#5$};
    }{
        \node (#1-label) at (#1.north) [anchor=south]{$#5$};
    }}}
    
}
\begin{tikzpicture}[node distance=2.5cm]
    \node (operation) [operation] {Operation};
    \drawmodule{iq}{above of=operation}{node distance=2cm}{i8}{i8}{e}{};
    \node (input) [io, above of=iq, node distance=1.4cm] {Inputs};
    \drawmodule{wq}{left of=operation}{rotate=90, anchor=center}{w8}{w8}{e-s}{};
    \node (weight) [io, left of=wq, rotate=90] {Weights};
    \drawmodule{bq}{right of=operation}{rotate=270, anchor=center}{b32}{b32}{w-s}{};
    \node (bias) [io, right of=bq, rotate=270] {Bias};
    \drawmodule{qout32}{below of=operation}{node distance=2cm}{o32}{o32}{e}{};
    \drawmodule{qout8}{below of=qout32}{node distance=2cm}{o8}{o8}{e}{};
    \node (output) [io, below of=qout8, node distance=1.4cm] {Outputs};
    \draw [arrow] (input) -- (iq);
    \draw [arrow] (iq) -- (operation);
    \draw [arrow] (wq) -- (operation);
    \draw [arrow] (weight) -- (wq);
    \draw [arrow] (bq) -- (operation);
    \draw [arrow] (bias) -- (bq);
    \draw [arrow] (operation) -- (qout32);
    \draw [arrow] (qout32) -- (qout8);
    \draw [arrow] (qout8) -- (output);
\end{tikzpicture}
}